\documentclass[runningheads]{llncs}
\usepackage{graphicx}

\usepackage{tikz}
\usepackage{comment}
\usepackage{amsmath,amssymb} 
\usepackage{wrapfig}
\usepackage{color}
\usepackage[section]{placeins}
\usepackage[accsupp]{axessibility}  

\usepackage[width=122mm,left=12mm,paperwidth=146mm,height=193mm,top=12mm,paperheight=217mm]{geometry}

\usepackage[pagebackref,breaklinks,colorlinks]{hyperref}
\usepackage{algorithm2e}
\usepackage{algpseudocode}
\usepackage{wrapfig}

\usepackage{overpic}
\usepackage{enumitem} 
\usepackage{overpic} 
\usepackage{color}

\definecolor{turquoise}{cmyk}{0.65,0,0.1,0.3}
\definecolor{purple}{rgb}{0.65,0,0.65}
\definecolor{dark_green}{rgb}{0, 0.5, 0}
\definecolor{orange}{rgb}{0.8, 0.6, 0.2}
\definecolor{red}{rgb}{0.8, 0.2, 0.2}
\definecolor{darkred}{rgb}{0.6, 0.1, 0.05}
\definecolor{blueish}{rgb}{0.0, 0.3, .6}
\definecolor{light_gray}{rgb}{0.7, 0.7, .7}
\definecolor{pink}{rgb}{1, 0, 1}
\definecolor{greyblue}{rgb}{0.25, 0.25, 1}

\usepackage{blindtext}

\usepackage{hyperref}

\renewcommand{\paragraph}[1]{\vspace{1em}\noindent\textbf{#1}.}

\begin{document}
\pagestyle{headings}
\mainmatter


\def\ECCVSubNumber{3778}  
\newcommand{\babak}[1]{{\texttt{\color{red} Babak: [{#1}]}}}
\newcommand{\kamran}[1]{{\texttt{\color{purple} Kamran: [{#1}]}}}
\title{Explaining Image Classifiers Using Contrastive Counterfactuals in Generative Latent Spaces} 

\titlerunning{Contrastive Counterfactuals in Generative Latent Spaces} 
\authorrunning{K. Alipour \emph{et al.}} 
\author{Kamran Alipour\inst{1}\and
Aditya Lahiri\inst{1} \and
Ehsan Adeli\inst{2} \and
Babak Salimi\inst{1} \and
Michael Pazzani\inst{1}}
\institute{UC San Diego \and Stanford University}

\maketitle

\begin{abstract}

Despite their high accuracies, modern complex image classifiers cannot be trusted for sensitive tasks due to their unknown decision-making process and potential biases. Counterfactual explanations are very effective in providing transparency for these black-box algorithms. Nevertheless, generating counterfactuals that can have a consistent impact on classifier outputs and yet expose interpretable feature changes is a very challenging task. We introduce a novel method to generate causal and yet interpretable counterfactual explanations for image classifiers using pretrained generative models without any re-training or conditioning. The generative models in this technique are not bound to be trained on the same data as the target classifier. We use this framework to obtain contrastive and causal sufficiency and necessity scores as global explanations for black-box classifiers. On the task of face attribute classification, we show how different attributes influence the classifier output by providing both causal and contrastive feature attributions, and the corresponding counterfactual images.

\end{abstract}
\section{Introduction}
\label{sec:intro}
Regardless of their accuracy, AI algorithms have yet to provide a level of interpretability to be accepted as trustable assets by their lay users in real-world applications. In recent years, eXplainable AI (XAI) has made an outstanding effort towards bringing transparency to AI with a focus on fairness and bias. Among different methods in this field, counterfactual explanations have received a lot of attention due to their scalable, intuitive, and logical approach~\cite{goyal2019counterfactual,mothilal2020explaining,wachter2017counterfactual,verma2020counterfactual}.

A counterfactual explanation for a black-box AI should provide transparency to the inner functionality of the algorithm through causal arguments and yet be interpretable to human users. Some methods specifically focus on the task of generating counterfactuals with a very high chance of changing AI's output such as adversarial examples~\cite{goodfellow2014explaining,guo2019simple}. On the other hand, studying the impact of interpretable attributes in the input on AI output usually goes as far as minimal correlations with model output and fails to meet the next two steps in Pearl’s ladder of causation~\cite{pearl2019seven}: intervention and counterfactuals.

\begin{figure}[t]
\begin{center}
\includegraphics[width=\linewidth]{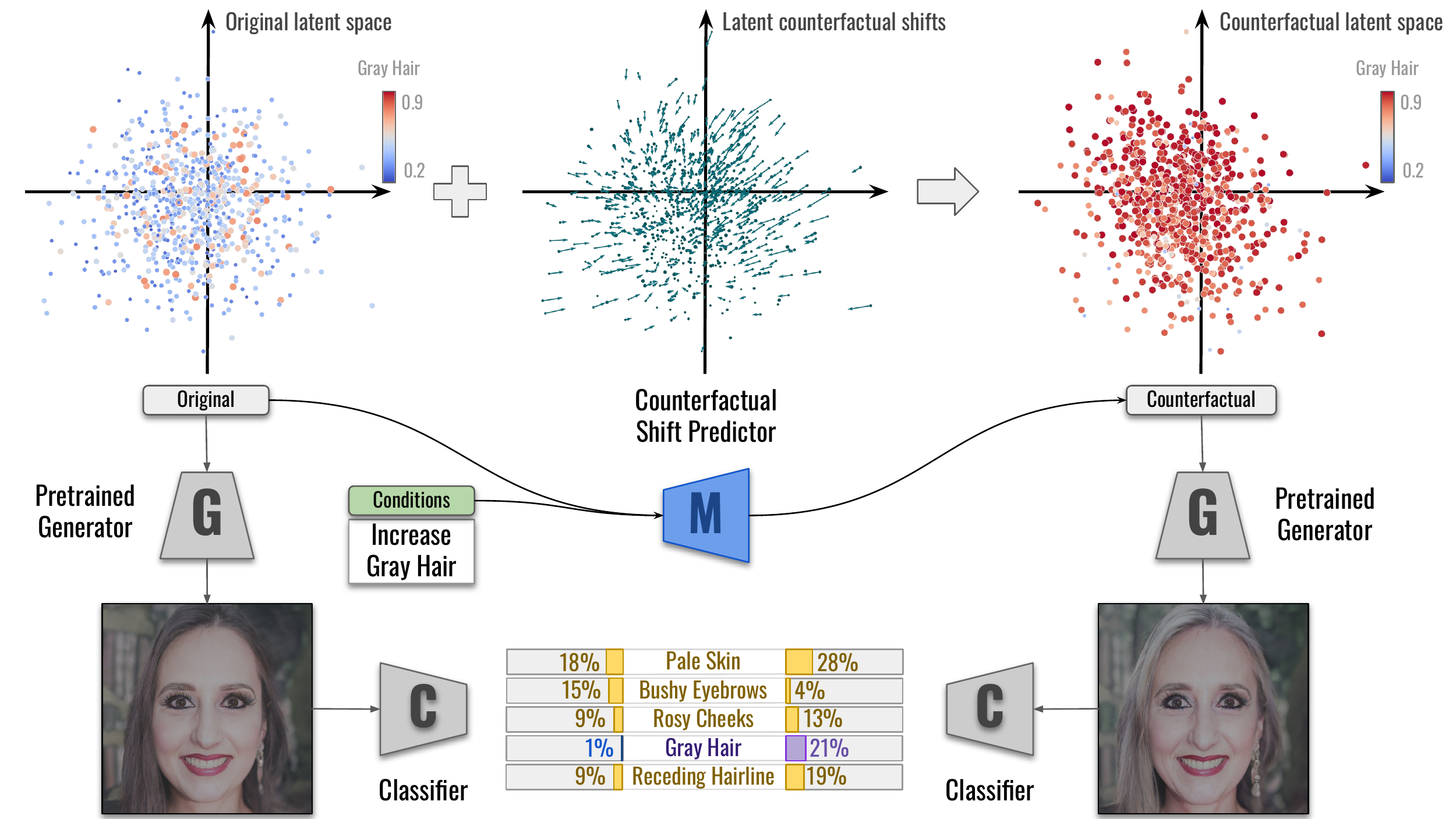}
\end{center}
\caption{
We introduce a method to learn counterfactual generation for a black-box classifier $C$. In this process, a shift-predictor $M$ is trained to predict contrastive counterfactuals for a classifier in the latent space of a generative model $G$. The shift-predictor is able to produce shift directions for any combination of attributes that are predicted by the classifier. Sampling from these probabilistic contrastive counterfactuals provides a framework to explain the biases and interactions across different predicted attributes.
}
\label{fig:teaser}
\end{figure}

While we acknowledge the fact that AI machines are not necessarily trained to follow causal reasoning based on interpretable features, we are interested in attributes that can provide the best of both worlds: to be as \textit{interpretable} and as \textit{causal} as possible. Learning such attributes can bridge the gap between causality and interpretability and lead to generating counterfactual explanations by changing meaningful attributes and still have a high causal influence on AI outcome.
Learning the interactions between causality and interpretability in feature space can bring us closer to the true definition of explainability of AI. Such a framework can also provide us with means to measure whether an AI machine is following any human-understandable pattern to produce its output or not.

Comparing and contrasting target points by observing their differences along a fixed set of understandable dimensions has been one of the primary ways in which humans have always explained and understood concepts~\cite{gerstenberg2015whether,morton2013contrastive}. These notions are also a  natural way of explaining image classifiers. We can see and understand the difference between a pair of images that are distinctly different from each other in certain known attributes. This enables us to reason about how those differences may cause them to obtain separate outcomes in some downstream tasks performed on them. Therefore, we aim to generate explanations of the following general form: ``For images with attributes having \(\langle\)value\(\rangle\) for which the algorithm made decision \(\langle\)outcome\(\rangle\), the decision would have been \(\langle\)foil-outcome\(\rangle\) with probability \(\langle\)score\(\rangle\) had the attribute been \(\langle\)counterfactual-value\(\rangle\)"~\cite{galhotra2021explaining}. In this work we seek to bring this contrastive framework for counterfactual explanations for image classification.\\
Notions of sufficiency and necessity build on this general form and allow us to reason about the necessary and sufficient conditions for a specific outcome. For an individual who received a positive outcome, necessity captures the importance of the existing value of an attribute in obtaining this outcome. On the other hand, for individuals who received a negative outcome, sufficiency reflects the ability of an attribute to flip the negative result into a positive one by modifying its existing value to some new value.
 Using a probabilistic interpretation of contrastive counterfactuals, we quantify the sufficiency and necessity of attributes to compute their causal responsibility towards the classifier's output. 
 
 Obtaining these probabilities of sufficiency and necessity is a challenging task. They require generating counterfactual images that reflect the exact changes we desire in a set of human-understandable attributes. In our work, we formalize the definition of these scores in the context of images and traverse the latent space of generative models to obtain these counterfactual images which correspond to user-defined values of the set of these interpretable attributes. This enables us to compute these scores efficiently.  
 An added benefit of our method is that instead of going through the expensive process of creating new datasets, it allows to sample a large number of inputs through pre-trained generative models and estimate contrastive counterfactuals over this sub-population for explaining any black-box image classification model (Fig \ref{fig:teaser}). 

In summary, the contributions of this work are as follows:
(1) We introduce a method to produce contrastive counterfactuals for an image classifier;
(2) Our method can use generative models pre-trained on any dataset and independent of the classifier training dataset;
(3) We propose contextual, contrastive, and causal explanations in the form of sufficiency and necessity scores to explain the black-box model.
(4) We use our method to provide global explanations for a black-box classifier trained on the CelebA dataset~\cite{liu2015faceattributes}. 


\section{Related works}
\label{sec:related}

Previous work in this area generally approaches the problem from several different perspectives. Some of the prior work take a fundamental approach and revolve around exposing the causal roots and achieving causal models.
A group of recent work take more rigorous approach by implementing contrastive counterfactuals for various applications. On the other hand, some of the methods in the vision area are centered around the use of generative models such as auto-encoders or GANs to produce interpretable counterfactuals. These generative approaches are divided to supervised and unsupervised techniques based on whether they involve annotations or classifiers in the process.

\paragraph{Causality}When estimating the causal effect of annotations, it's important to consider the confoundedness between these attributes for the purpose of any intervention. In that regard, most of the previous work attempts to learn a form of a structural causal model (SCM)~\cite{sani2020explaining,pawlowski2020deep}.
 Parafita \emph{et al.}~\cite{parafita2019explaining} use causal counterfactuals to provide explanations by obtaining attributions for known latent factors. Dash \emph{et al.}~\cite{dash2020evaluating} use a conditional GAN to generate counterfactuals. Bahadori \emph{et al.}~\cite{bahadori2020debiasing} use a causal prior graph and existing annotations to explain the predictions. 
Khademi and Honavar~\cite{khademi2020causal} compare different methods of causal effect estimation such as CBPS and NPCBPS to interpret predictive models and explain their prediction based on inputs average causal effect (ACE). In a different approach, Zaeem and Komeili~\cite{zaeem2021cause} introduce interpretable attributes as “concepts” and propose learning the presence of each “concept” in different layers of the classifier. Ghorbani \emph{et al.}~\cite{ghorbani2019towards} also develop a systemic framework to automatically identify higher-level concepts which are both human-interpretable and important for the ML model. However, these concepts are neither necessarily causal nor require any prior interpretable attributes as input. While these techniques provide comprehensive explanations on the causal effect of attributes on model output, yet the causality is often quantified over a population and in correlation metrics. Such correlations are represented as global explanations and satisfy the first step of Pearl’s ladder of causation~\cite{pearl2019seven}, however, they cannot guarantee a causal impact on a case-by-case intervention (local explanation). In this work, we aim to go beyond correlations and provide a framework for a complete implementation of the causation ladder.

\paragraph{Contrastive counterfactuals}
Contrastive counterfactuals have been the building blocks of ideas in philosophy and cognition that guide people's understanding and dictate how we explain things to one another~\cite{de2017people,morton2013contrastive}, and have been argued to be central to explainable AI~\cite{miller2019explanation}. 
To quantify these notions, probabilistic measures have been formalized and applied to a variety of fields including AI, epistemology, and legal reasoning~\cite{russell2017worlds,greenland1999epidemiology}.
Recent work has also focused on using them in the field of Explainable AI~\cite{galhotra2021explaining,kommiya2021towards,watson2021local}.
Our work is following a trend of ongoing research into generating counterfactual explanations for AI algorithms \cite{bertossi2020causality,karimi2020model,laugel2017inverse,mahajan2019preserving,mothilal2020explaining,ustun2019actionable,verma2020counterfactual,wachter2017counterfactual,stepin2021survey}. We are specifically interested in the implementation of this framework in the image classification problem. This topic is inherently challenging as it demands a probabilistic causal model based on the algorithm's output.

\paragraph{Explainable Autoencoders} Due to their strong abilities in representation learning, auto-encoders have received a lot of attention in the XAI community. Variational auto-encoders (VAEs) have shown promising results in learning causal~\cite{o2020generative} and interpretable~\cite{an2021exon} representations or interception of interpretable attributes~\cite{goyal2019explaining}. Similarly, Castro \emph{et al.}~\cite{castro2019morpho} propose a framework to measure how much the latent features in a VAE represent the morphometric attributes in the MNIST images. Some studies propose a feature importance estimation based on Granger causality~\cite{thiagarajan2020accurate,schwab2019cxplain}. While auto-encoders are very strong in representation learning, they tend to lose detail in regenerating highly complex data. We focus our approach on the use of pre-trained generative models that can produce high-resolution counterfactuals with accurate shifts in interpretable features.

\paragraph{Unsupervised disentanglement in GANs} Within recent related work in this area, a number of them are dedicated to unsupervised disentanglement of latent space of GAN models to interpretable feature spaces. Some approaches use fundamental techniques such as projection~\cite{shen2020interpreting}, PCA~\cite{harkonen2020ganspace}, and orthogonal regularization~\cite{liu2022towards}, while others use self-supervised techniques to learn interpretable representations~\cite{nitzan2020face}. Another popular approach is the unsupervised discovery of linear~\cite{lu2020unsupervised,voynov2020unsupervised} or nonlinear~\cite{tzelepis2021warpedganspace} directions that correlate with interpretable features. Moreover, adversarial methods~\cite{yang2021discovering} alongside contrastive learning-based and intervention-based approaches~\cite{yuksel2021latentclr,gat2021latent} have also been studied for the purpose of interpretable direction discovering for GANs. 
Despite their impressive results, these techniques may combine multiple distinguishable attributes in one detected direction. On the other hand, the detected interpretable directions are not always easy to label and hence cannot be used for any label correction or model improvement. In this work, we propose learning latent directions that correspond to actual labels so the explanation results can be used in improving datasets and training procedures.

\paragraph{Supervised direction discovery}On the other hand, a large number of contributions in this area are focused on supervised discovery of interpretable directions in the latent space of generative models. The majority of these models are implemented for GANs such as StyleGAN~\cite{karras2020analyzing} due to their resounding success and high quality. A classic solution in this area is finding the class boundary hyper-plane in the latent space of GAN~\cite{li2021discover}. There are approaches that attempt to find counterfactuals with the use of gradient descent in a GAN's latent space~\cite{liu2019generative}. Some of the existing approaches train GANs to either apply residuals~\cite{nemirovsky2020countergan} or masked transformations~\cite{samangouei2018explaingan} on images for the purpose of counterfactual generation. Moreover, some prior work experiment with incorporating the classifier~\cite{lang2021explaining} or contrastive language-image models~\cite{patashnik2021styleclip} into GAN to accommodate attributes into the latent space. Another novel approach is the use of energy-based models (EBMs) for a controllable generation with GAN, however, this technique requires manual labeling of the latent samples \cite{nie2021controllable}. Styleflow~\cite{abdal2021styleflow} introduces counterfactuals with conditional continuous normalizing flows in the latent space, however, their solution is specifically tailored for the extended latent space of StyleGAN. In our proposed approach, we seek a minimal training process by utilizing only pretrained GANs. Our methodology follows a simple and scalable implementation to be compatible with different generative models.
\section{Method}
\begin{figure*}[t]
\begin{center}
\includegraphics[width=.99\columnwidth]{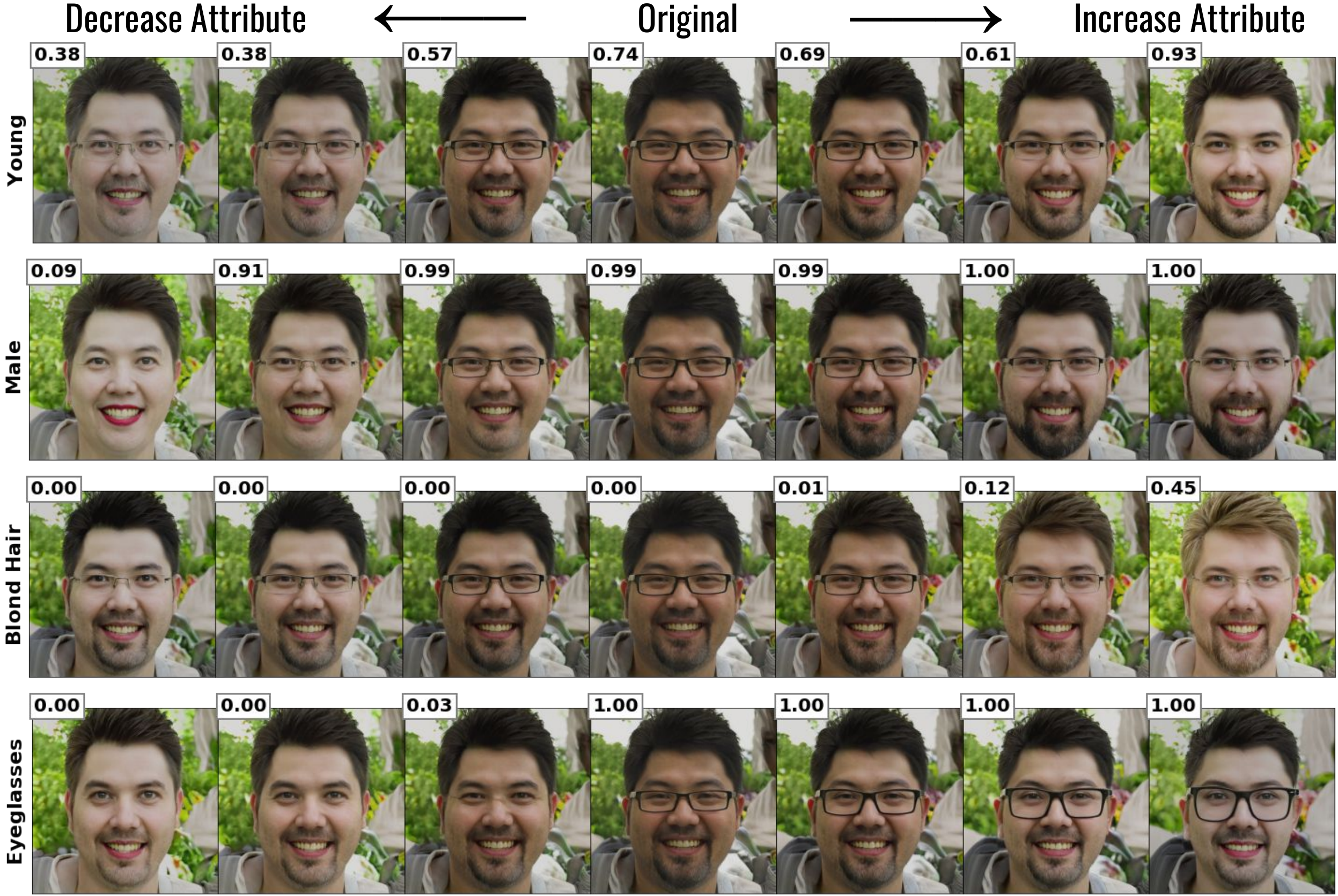}
\end{center}
\caption{
A shift predictor learns to predict optimum shifts in proximity of any input and manipulate the input for different attributes and different directions. Aside from the intended attribute that has changed in each direction, some other attributes also change which shows the shift predictor's ability to sample from a more realistic distribution and be mindful of potential confoundedness across different attributes.
}
\label{fig:bidirectional_CFs}
\end{figure*}

In this section, we first discuss the kind of explanations that can be obtained using  contrastive counterfactuals and provide some necessary background. Following that, we formalize the probability of sufficiency and necessity, which are at the core of our explanations. We also describe how we can compute those scores in the setting of image classifiers.
We later explain the algorithm pipeline, which consists of a pretrained generative model \(G\) to produce realistic images and introduce a shift predictor to achieve counterfactual latent vectors for \(G\) (Fig \ref{fig:bidirectional_CFs}). These allow us to compute the necessity and sufficiency scores for explaining black-box image classifiers.

\subsection{Contrastive Counterfactual Explanations}
\label{sec:method_cc}
Our goal is to use probabilistic contrastive counterfactuals and develop a feature attribution method that generates explanation for an image classifier. These feature attributions quantify the causal contribution of a set of interpretable attributes on the outcome of the classifier. Specifically, for an image classifier which predicts the output \(Y\) for input images with an interpretable attribute \(A\), our framework generates explanations of the following form: ``For an input image with an attribute \(a\) for which the classifier outcome is \(y\), the classifier outcome would be \(\hat{y}\) with  probability \(s\), had the input attribute been \(\hat{a}\) instead of \(a\)". In the task of attractiveness classification for face images, these explanations pertain to images that had the positive outcome of being classified as attractive. For those cases, such explanations measure the probability with which increasing an interpretable attribute such as baldness could lead to a negative outcome instead. 
Therefore, they measure the extent to which the original value of the attribute is necessary for positive outcomes, hence called {\em probability of necessity}. Complementary to that we provide sufficiency scores for input images that receive the negative output. Such explanations compute the extent to which changing an attribute is sufficient to flip a negative outcome to a positive one, hence called {\em probability of sufficiency}. 

In this work, we rely on Pearl's probabilistic causal models~\cite{pearl2009causality} to formalize and evaluate the notions of probability of sufficiency and necessity. Next, we briefly review probabilistic causal models and then build on that to mathematically define \textit{Necessity} and \textit{Sufficiency} scores.





\par {\bf Causal models and counterfactuals.}
A probabilistic causal model (PCM) consists of (1) a set of observable \emph{endogenous} attributes $\mathcal{A}$, (2) a set of latent \emph{background or exogenous} variables $\mathcal{U}$, (3) a set of structural equations $\mathcal{F}$ that capture the causal dependencies between the attributes by associating a function $F_A \in \mathcal{F}$ to each endogenous attribute $A \in \mathcal{A}$ that expresses the values of each endogenous attribute in terms of $\mathcal{U}$ and $\mathcal{A}$, and (4) a probability distribution $P(\mathbf{u})$ over the exogenous variables $\mathcal{U}$.
Given a probabilistic causal model, an intervention on an endogenous attribute $A \subseteq \mathcal{A}$, denoted 
$A \leftarrow {a}$, is an operation that modifies the underlying causal model by replacing $F_A$, the structural equations associated with $A$, with a constant $a$. The \emph{potential outcome} of an attribute $Y$ after the intervention $A \leftarrow a$ in a context of exogenous variables $u$, denoted $Y_{A \leftarrow a}(u)$, is the solution to $Y$ in the modified set of structural equations. 

The distribution $P(\mathbf{u})$ induces a probability distribution over endogenous attributes and potential outcomes. Considering proper PCMs, one can express counterfactual queries of the form $P(Y_{ A \leftarrow a}=\hat{y})$, or simply $P(\hat{y}_{ A \leftarrow a})$; this reads as ``What is the probability that we would observe $Y=\hat{y}$ had $A$ been $a$?" and is given by the following expression: 
%
%
 \begin{equation} \small
 \begin{aligned}
  P(\hat{y}_{ A \leftarrow a}) = \sum_{u} \ P(\hat{y}_{ A \leftarrow  a}(u)) \ P(u). 
  \end{aligned}
  \label{eq:prob_contrastive_CF}
\end{equation}
\par {\bf Probability of Necessity and Sufficiency.} 
We are given a binary image classifier with the output $Y=\{y,\hat{y}\}$, where $y=1.0$ and $\hat{y}=0.0$ denote the positive (favorable) and negative (unfavorable) outputs respectively, and a binary attribute $A=\{a,\hat{a}\}$ associated with the input images, where $a=1.0$ and $\hat{a}=0.0$ respectively denote the presence or absence of the attribute.
The probability of SUFficiency and NECessity of $A$ for $Y$ measures as follows:\\
 \begin{equation}
   NEC = P(\hat{y} _{A\leftarrow \hat{a}}\mid a,y),
  \label{eq:Nec}
\end{equation}
 \begin{equation}
   SUF = P(y _{A\leftarrow a}\mid \hat{a},\hat{y}).
  \label{eq:SUF}
\end{equation}
%
%
Given a sub-population of input images with attributes $a$, for which the classifier returns the positive output, the notion of probability of necessity~\eqref{eq:Nec} captures the probability that on changing the attribute $A$ from its default value of $a$ to the intervened value of $\hat{a}$, the classifier will return a negative outcome instead. In other words, it 
measures the extent of positive classifications that are attributable to the {\em original state} of the attribute $A=a$. The probability of sufficiency~\eqref{eq:SUF} is the dual of the probability of necessity. It applies to sub-population of input images with default attribute value $\hat{a}$, for which the classifier produced a negative output. It measures the effect of changing the attribute by intervention to $a$ from its default state of  $\hat{a}$. It computes the probability that this change could cause the classifier to return a positive outcome for these cases which were originally handed out a negative outcome. Hence, it measures the {\em capacity} of setting $A$ to $a$ to {\em flip} the negative outcome from the classifier.

We can choose to \emph{change} the attribute from its default state by moving in its direction of increase or decrease. This would allow us to measure the sufficiency and necessity of changing the attributes in both directions and give a more in-depth understanding of how features are influencing the outcome of the black-box classifier. We denote necessity scores of increasing the attribute with $NEC^{+}$ and the necessity scores of decreasing the attribute with $NEC^{-}$. We follow a similar notation for sufficiency scores. 

\par {\bf Computing Necessity and Sufficiency.}  

\setlength{\intextsep}{0pt}%
\setlength{\columnsep}{1pt}%
\begin{wrapfigure}{r}{0.3\textwidth}
\centering
\includegraphics[width=.2\columnwidth]{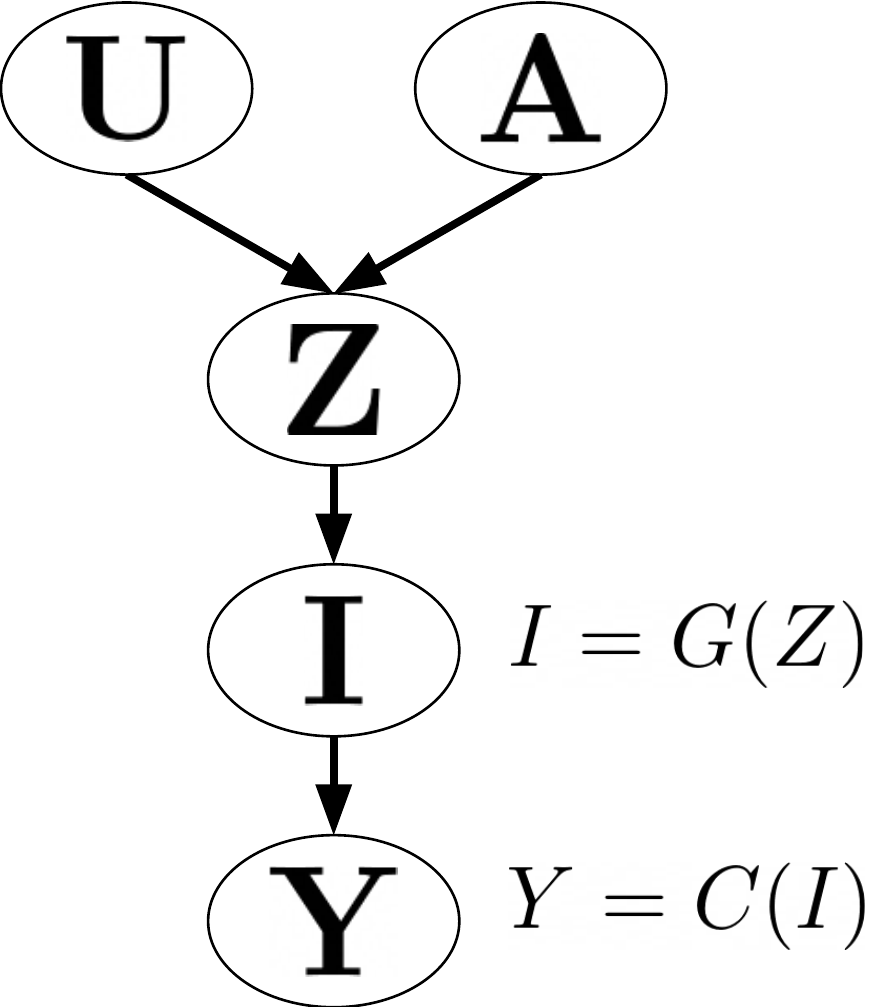}
\caption{Causal model.}
\label{fig:causal_model}
\end{wrapfigure}

\noindent We assume a causal model as shown in Fig. \ref{fig:causal_model} that has the following components.   \emph{a}) Unobserved attributes ($\mathcal{U}$) - These are all the attributes that we do not have any observations for, and  cannot account for. They constitute the exogenous variable in our causal model. \emph{b}) A set of  interpretable attributes ($\mathcal{A}$)- These are a set of attributes that we have known values for. They are our endogenous variables in the causal model. \emph{c}) A latent space (\textbf{Z}) - The unobserved (exogenous) attributes, and observed set of interpretable attributes (endogenous) together directly affect the value of the latent space. \emph{d}) A generative model  (\(G\)) - It takes as input the above mentioned latent space \textbf{Z} and transforms it to an image $I$. \emph{e}) The classifier to be explained ($C$) - It takes as input the image $I$ and produces the target label, $Y$.

We generate counterfactuals, and also pass the produced images (both originals and counterfactuals) to the black-box classifier to obtain the classifier's output. We use this information to compute the sufficiency and necessity scores. We generate counterfactuals by following Pearl's three-step procedure~\cite{pearl2009causality}. 
\begin{itemize}
    
    \item \textbf{Abduction} Given the prior distribution of the latent variable of a generative model \(G\) and the set of attributes \(\mathcal{A}\), train a model \(M\) that estimates the updated probability of latent variable conditioned on any subset of  attributes: 
    \begin{equation}
      P(\textbf{z} \mid \bar{\mathcal{A}}\ = \textbf{a}),\ \bar{\mathcal{A}} \subseteq \mathcal{A}.
      \label{eq:abduction}
    \end{equation}
    \item \textbf{Action} Take a set of interpretable attributes $\bar{\mathcal{A}}$. Based on the causal model, perform an intervention by setting a subset of attributes $\bar{\mathcal{A}} \subseteq \mathcal{A}$ to their determined values \(\bar{\mathcal{A}} \leftarrow \hat{\textbf{a}}\).
    \item \textbf{Prediction} Given the model \(M\), obtain the modified latent vector probability that corresponds to the new value of the attribute(s). Pass this modified latent vector into the generative model to obtain the corresponding image. Finally, run black-box classifier inference for this counterfactual image to obtain target output:
    \begin{equation}
    \begin{gathered}
       \hat{I} = G(\hat{\mathbf{z}}),\  \hat{\mathbf{z}} = M(\textbf{z}, \bar{\mathcal{A}}=\hat{\textbf{a}}), \\
      \hat{Y} = C(\hat{I}).
    \end{gathered}
      \label{eq:prediction}
    \end{equation}

\end{itemize}

\subsection{Counterfactual Generation}

Conducting the three steps of counterfactual generation is a non-trivial task due to the complication that arises when setting the attributes in the Abduction and Prediction steps. Specifically, generative models tend to have very complex latent spaces, where finding a path from \(\textbf{z}\) to \(\hat{\textbf{z}}\) for the purpose of attribute change is intractable. To reconcile for it, we propose training a MLP model \(M\) that serves as the shift predictor in our pipeline and can provide prediction for \(\hat{\mathbf{z}} =M(\textbf{z} , \bar{\mathcal{A}}=\textbf{a})\). With the use of \(M\), we can now update the probability of latent variable \(P(\textbf{z})\) to the probability of counterfactual latent variable \(P(\textbf{z} \mid \bar{\mathcal{A}}=\textbf{a})\) in the prediction step and follow Pearl's procedure. In the following, we provide the details on the generative model and shift predictor algorithm.
\begin{figure}[t]
\centering
\includegraphics[width=.99\columnwidth]{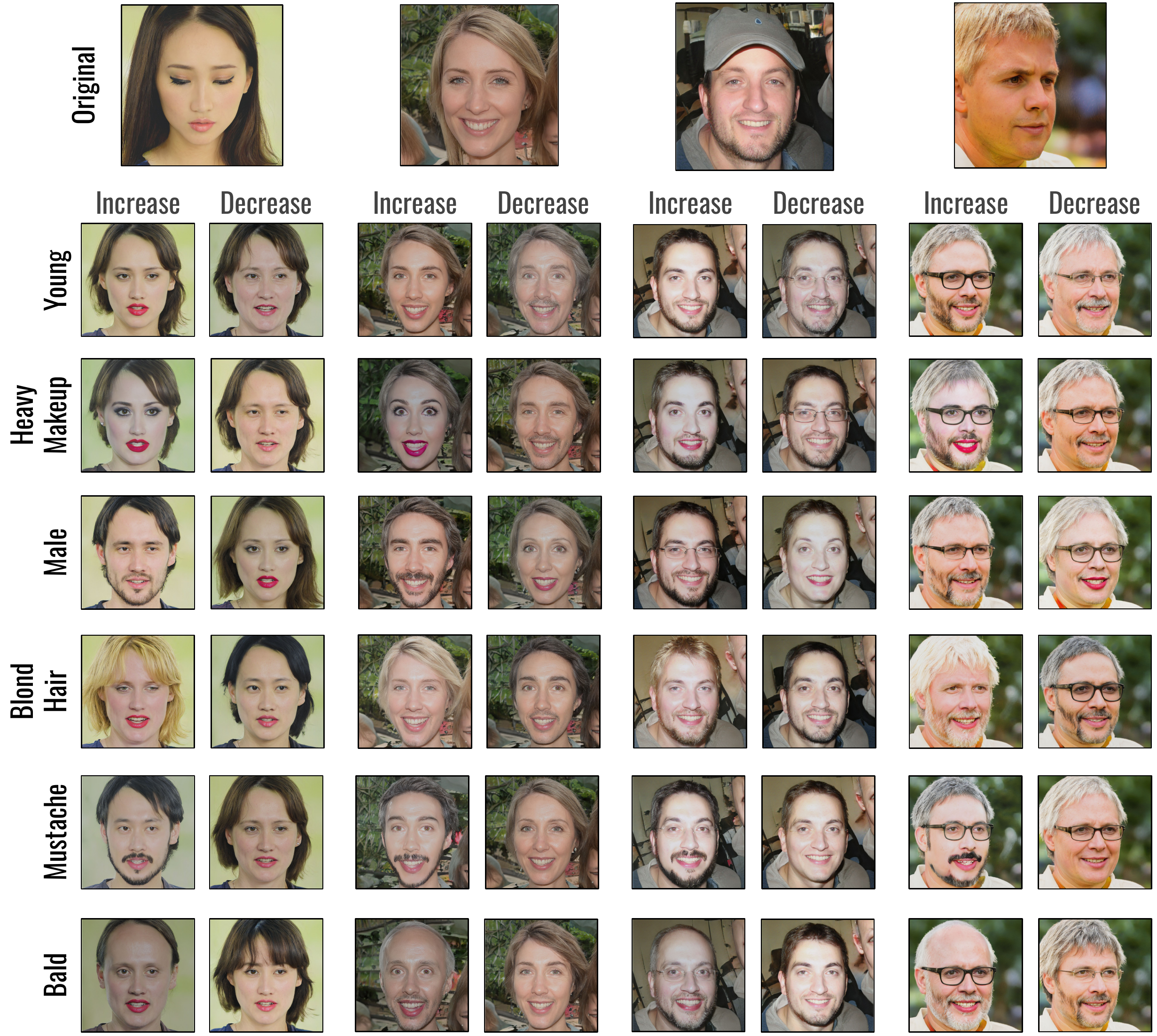}
\caption{Examples of counterfactual images that are generated during computation of the explanations scores. First row shows the original images. Each subsequent row shows the original image being modified to move towards the direction of increase and decrease of the labelled attribute. Along with necessity and sufficiency scores, these representative images are provided to the user as global explanations for the classifier.}
\label{fig:global_images}
\end{figure}
\par {\bf Generative model.}
Generative models are vastly popular in different fields of AI, and their recent advancements in creating realistic images have made them a viable approach to producing a latent representation of an image dataset. In our experiments, we utilize StyleGAN2~\cite{karras2020analyzing} as a state-of-the-art generative model which can be used to generate high resolution and realistic images in different domains. StyleGAN feeds the latent variable into a mapping network that transforms it into an intermediate latent variable. Aside from its ability to produce styles, this transformation also provides the intermediary latent space as a more regulated domain to learn and traverse through interpretable attributes.

\par  {\bf Shift predictor.} A shift predictor model is an MLP model that can take the latent variable of an image from a generative model \(G\) and generates the latent variable for its counterfactual based on the attributes produced by a classifier (Fig. \ref{fig:global_images}). For a generative model \(G: \mathcal{R}^d\rightarrow \mathcal{R}^n\) that has a latent space with dimension \(d\) and a classifier \(C: \mathcal{R}^n \rightarrow \mathcal{R}^m\) that predicts \(m\) attributes, we define our shift predictor as \(M(\textbf{z},\hat{\textbf{y}}): \mathcal{R}^d \times \mathcal{R}^m\rightarrow \mathcal{R}^d\), where  \(\textbf{z} \in \mathcal{R}^d\) is the latent variable for the input image and \(\hat{\textbf{y}}\) denotes the attributes for the intended counterfactuals. In the training process, shift predictor learns the directions in the latent space of \(G\) that correspond to changes in the attributes predicted by the classifier. Without the need for any manual labeling, the training procedure only requires the latent variables of images from \(G\) to input the shift predictor and supervise it with the labels generated by the classifier (see Alg. \ref{alg:trainpredictor}).

\RestyleAlgo{ruled}
\SetKwProg{Init}{init}{}{}
\begin{algorithm}[t] \small
\DontPrintSemicolon
\caption{Training a shift predictor for binary attributes}\label{alg:trainpredictor}
\KwData{Classifier \(C\). GAN model \(G\). Counterfactual faithfulness ratio \(\gamma\).}
\KwResult{Parameters $\Theta_M$ for the shift predictor model \(M\).}
\Init{:}{
$\Theta_M \gets $ Random initialization\;
}
\For{number of iterations}{
Sample a batch of \(b\) noise variables and target outputs:\;
\( \{ \textbf{z}^{(1)}, ... , \textbf{z}^{(b)}\} \gets p(\textbf{z}) \sim \mathcal{N}(\mathbf{0},I)\)\;
\(\{ \hat{\textbf{y}}_{i}^{(1)}, ... , \hat{\textbf{y}}_{i}^{(b)}\} \leftarrow Bern.(p=0.5),\ \forall i=1 .. m\)\;
Predict counterfactual latent codes:\;
\( \hat{\textbf{z}}^{(j)} \gets M(\textbf{z}^{(j)},\hat{\textbf{y}}^{(j)}),\ \forall j=1 .. b\)\;
Generate images from the noise variables and predict attributes by the classifier:\;
\(I^{(j)} \gets G(\hat{\textbf{z}}^{(j)}),\ \forall j=1 .. b\)\;
\(\textbf{y}^{(j)} \gets C(I^{(j)}),\ \forall j=1 .. b\)\;
Compute attribute conditioning and faithfulness loss:\;
\(\mathcal{L}_{a} \gets \frac{1}{b}\sum_{j=1}^{b}\sum_{i=1}^{m}-\hat{\textbf{y}}_i^{(j)}log(\textbf{y}_i^{(j)})\)\;
\(\mathcal{L}_{f} \gets \frac{1}{b}\sum_{j=1}^{b}||\hat{\textbf{z}}^{(j)} - \textbf{z}^{(j)}||\)\;
Update the shift predictor parameters:
\(M \gets \triangledown\Theta_M (\mathcal{L}_a+\gamma\mathcal{L}_f)\)\;
}
\end{algorithm}

During the training, shift predictor learns to produce a counterfactual latent variable that satisfies any combination of attributes defined by \(\hat{\textbf{y}}\). In other words, if the classifier predicts a set of attributes \(\mathcal{A} = \{A_1, A_2, ..., A_m\}\), shift predictor can provide a counterfactual latent variable compatible with any selected subset of attributes \(\bar{\mathcal{A}}\):
\begin{equation}
  \hat{\textbf{z}} = M(\textbf{z},\{A_i=\hat{a}_i\ | A_i \in \bar{\mathcal{A}}\} ).
  \label{eq:ShiftPredictorInference}
\end{equation}
Under the assumption of proper training, the shift predictor is an approximation of latent variable distribution conditioned by the subset of attributes \(\bar{\mathcal{A}}\) :
\begin{equation}
  \hat{\textbf{z}} \sim P(\textbf{z}|\{A_i=\hat{a}_i\ | A_i \in \bar{\mathcal{A}}\} ),\ \bar{\mathcal{A}} \subseteq \mathcal{A}.
  \label{eq:zhatDistribution}
\end{equation}
 The loss function in the training process pursues two objectives: 1) minimizing the error in prediction of attributes \(\bar{\mathcal{A}}\) for the counterfactual image, 2) assuring a level of faithfulness and similarity between to the original input image and its newly generated counterfactual. The attribute loss \(\mathcal{L}_{a}\) is defined as a cross entropy between the conditioned attributes and the attributes predicted by the classifier. In the training process, the conditioned attributes \(\bar{\mathcal{A}}\) are distinguished from unset attributes so the loss will be only calculated for them. On the other hand, the faithfulness loss \(\mathcal{L}_{f} \) is calculated as the normal distance between the original latent variables and their counterfactuals. The overall loss in the training process is defined as a combination of these two losses with a faithfulness factor \(\gamma\) which defines a balance between attribute accuracy of counterfactuals and their faithfulness to the original input:
 \begin{equation}
   \mathcal{L} = \mathcal{L}_a + \gamma \mathcal{L}_f = \sum_{A_i \in \bar{\mathcal{A}}}
   -\hat{\textbf{y}}_i log(\textbf{y}_i) + \gamma ||\hat{\textbf{z}} - \textbf{z}||
  \label{eq:FullLoss}
\end{equation}


\section{Experiments and Results}

We run our experiments to explain black-box classifiers that are trained on the task of classifying face images.  We have annotations for the set of interpretable attributes \(A\) that we will choose to use to explain the model's behavior. We train a multi-task classifier built on top of a pretrained VGG~\cite{simonyan2014very} backbone to predict the set of interpretable attributes \(A\) for new unseen images. The CelebA dataset is used as the training set and provides 39 binary attributes including attractiveness which we use as the target output \(Y\) for any black-box classifier of choice. As the set of explanatory interpretable attributes(\(A\)), we choose six other labels: blonde hair, heavy makeup, baldness, mustache, youngness, and maleness. We make a simplifying assumption to use an underlying causal model in which the explanatory attributes are independent of each other. We model attractiveness as the positive class $(y = 1.0)$ and unattractiveness as the negative class $(\hat{y} = 0.0)$ for the black-box classifier to predict. In the set of interpretable attributes(\(A\)), an attribute has its default value as $(a)$ when it is not explicitly set. Otherwise, during intervention, it is set to value ($\hat{a}$) which can be $+1$ if we want to move in the direction of its increase, and $-1$ if we want to move in the direction of its decrease. We intervene and set $\hat{a}$ to $0$ if we do not wish to modify the attribute. Our initial dataset consists of 200 images randomly produced by the generative model. We pass the images through the multi-task classifier to obtain the attribute values and through the black-box classifier to obtain the target label. We seek to explain the behavior of the target output \(Y\) using the attributes \(A\) on this dataset by performing the following experiments:
\begin{itemize}

  \item \textbf{Linear baseline}, we first consider an interpretable linear approximation of target label behavior w.r.t. the attributes as the underlying black-box model. We use this approximation as ground truth and assess the validity of necessity and sufficiency scores in capturing this ground truth linear behavior.
  \item \textbf{Black-box explanations}, we consider a complex black-box classifier built on a pretrained VGG
backbone and generate sufficiency and necessity scores to explain it. In conjunction with generated counterfactual images, we use these scores to analyze how increasing and decreasing the attributes affects the classification into attractive and not-attractive labels by the classifier.
\end{itemize}

\begin{figure}[t]
\centering
\includegraphics[width=.99\columnwidth]{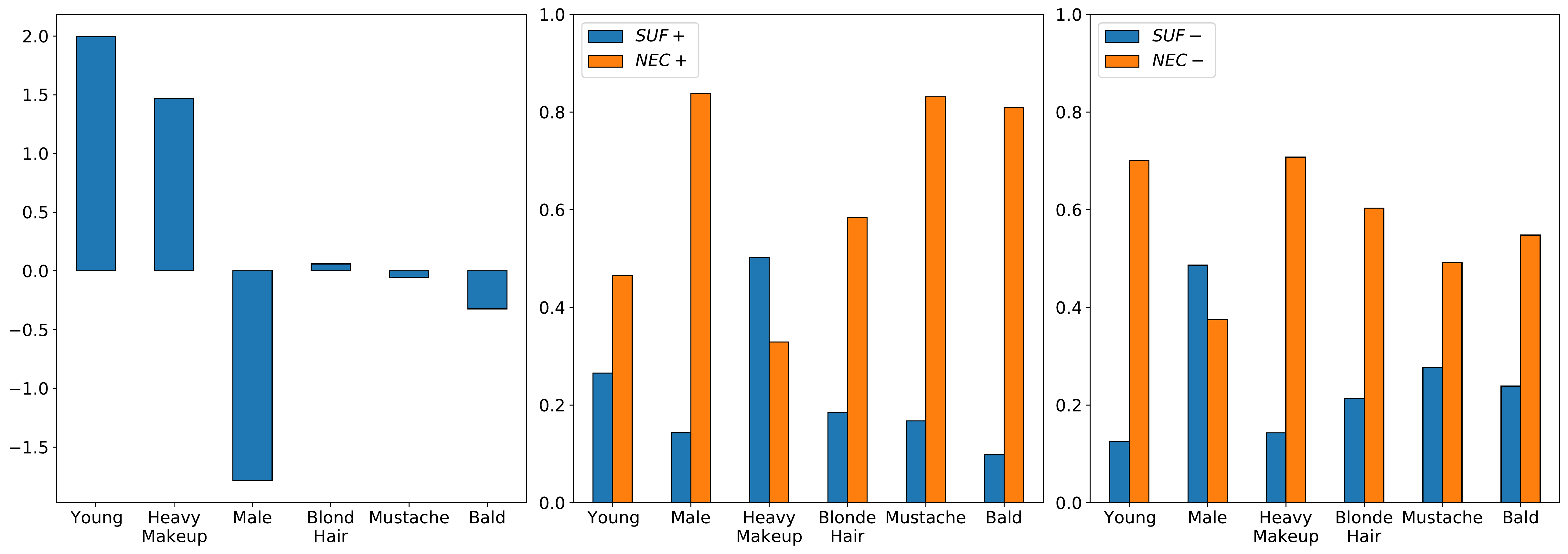}
\caption{\textbf{Left}: The coefficients of the known black-box logistic regressor. \textbf{Center}: The sufficiency and necessity scores explaining the logistic regressor behavior when attributes are increased. \textbf{Right}: The sufficiency and necessity scores explaining the logistic regressor behavior when attributes are decreased.}
\label{fig:log_reg_compare}
\end{figure}

\subsection{Linear baseline}
Our pipeline of generating counterfactual images and explanation scores is agnostic to the type of black-box model being explained. This implies that it is independent of the type of machine learning or deep learning model used. However, one way to test the quality of our explanations is by generating explanations for a case where we actually have access to the decision making rationale of the underlying black-box model. To this end, we choose a logistic regression classifier as the model whose decision we seek to explain. This classifier takes real values corresponding to the feature attributes obtained from the multi-task classifier, and predicts the target label of attractiveness using only these values. The coefficients of the logistic regressor corresponding to the different features gives us an indication of how the model is making its decisions. We compare this to sufficiency and necessity scores generated by our method which seeks to explain this logistic regression model. \\
We observe from Fig. \ref{fig:log_reg_compare} that indeed, features that have negative attributions such as those for male, mustache and bald, in the logistic regressor model carry a large value of necessity when we move in directions of their increase. This high necessity score indicates that leaving them unset as compared to increasing them is most important to allow attractive individuals keep their attractiveness score high. Similarly, for people classified as unattractive, the features that had high positive attributions in the logistic regressor such as Young and Heavy Makeup, carry the highest values of sufficiency when we increase their values. This is indicative of the fact that in order to flip the outcome from not-attractive to attractive, increasing heavy makeup and youngness are the two most important factors. We can interpret the $SUF/NEC^{-}$ scores in a similar way. This kind of contrastive and counterfactual analysis is not possible through simple coefficients obtained from the logistic regressor. This makes it important to use these notions of sufficiency and necessity over standard co-efficient based attributions that have been observed to have multiple shortcomings due to their overly simplistic nature. These include issues like their dependence on feature pre-processing methods, as well as instability due to different feature selection\cite{lipton2018mythos}.

\subsection{Black-box Explanations}

We use our pipeline to generate global explanations for the black-box attractiveness classifier. Here, the black-box classifier is built on a pretrained VGG backbone. Our explanations are two-fold. First, we provide sufficiency and necessity scores on a population level for the 6 different attributes. In addition to this, we also provide users with the counterfactual images that were generated by our shift predictor during the computation of the scores. The ability to have both feature attributions, as well as the images that led to the computation of those attributions allows the user to understand model behavior at a deeper level.

\begin{figure}[t]
\centering

\includegraphics[width=.99\columnwidth]{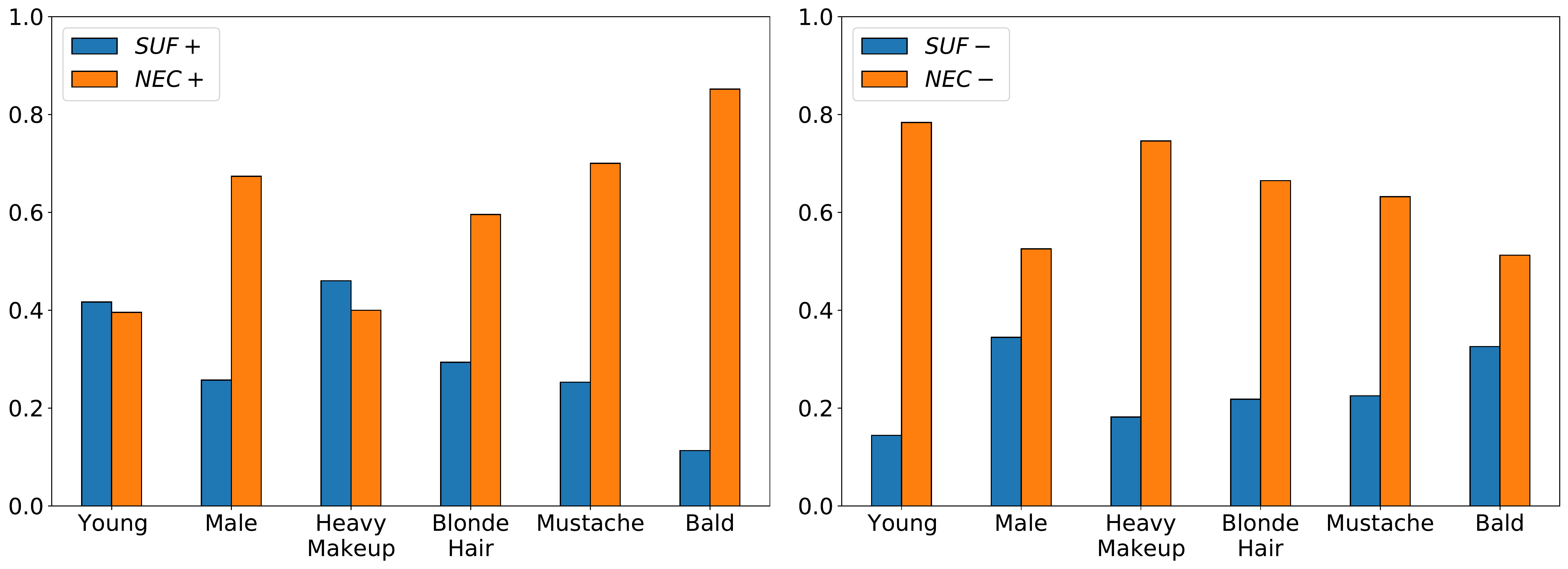}
\caption{Sufficiency and necessity scores as global explanations. \textbf{Left:} Not increasing baldness and mustache are most necessary to remain classified as attractive. Increasing heavy makeup and youngness is most sufficient to flip an unattractive outcome to attractive.  \textbf{Right:} Not decreasing youngness and heavy makeup is most necessary to stay classified as attractive. Decreasing maleness and baldness are most sufficient to reverse the outcome from unattractive to attractive. }
\label{fig:global_scores}
\end{figure}

Fig. \ref{fig:global_images} contains a set of representative images. We can see how the original image changes in the direction of decrease and increase of the explanatory attributes. Fig. \ref{fig:global_scores} shows the overall sufficiency and necessity scores of attributes in both positive and negative directions of increase. This gives us a detailed analysis of how the features are affecting the classifier output. For instance, a high $SUF^+$ value of the attributes Heavy Makeup, Blond Hair, and Young implies that moving in the direction of increase of these attributes is most sufficient to flip an outcome of unattractiveness to attractive. Similarly, a high $SUF^-$ value of the attributes Male, and Bald reflects that when moving in the direction of decrease of these attributes, we are most likely to be able to flip our outcome from not attractive to attractive. The necessity scores inform us about the attributes that are most important to be left "unset" in their default state as compared to increasing or decreasing them, in order for a person classified as attractive to maintain that classification. A high $NEC^+$ score of Baldness, Moustache, and Male is indicative of the fact that one should avoid increasing these attributes if they wish to remain classified as attractive by the classifier. Similarly, the high $NEC^-$ scores of Young, Heavy Makeup, and Blond Hair are indicative of the fact that one should avoid decreasing these attributes if they wish to remain classified as attractive by the classifier. With the generated counterfactuals images as evidence, the necessity and sufficiency scores provide a holistic understanding of the black-box classifier to the end-user.

\section{Conclusions}
In this work, we proposed an end-to-end pipeline that generated counterfactuals from a pretrained generative model and used that to help compute probabilistic causal counterfactual scores. These scores, along with the generated images, served as explanations for any underlying black-box image classifier. Our work also highlighted the need and advantages of these contrastive explanations over simple feature attributions. However, one of the drawbacks of our current method is that it does not effectively disentangle the effects between attributes. We would want to improve on that aspect by learning a structural causal model that can model the effects which attributes have on one another as well. This would also allow us to extend our analysis to compute the direct and indirect effects \cite{pearl2013direct} of attributes on the target label. Furthermore, we would like to apply this pipeline to detect and mitigate bias in image classification systems.

\clearpage
%
%
\bibliographystyle{splncs04}
\bibliography{egbib}

\end{document}